\documentclass[letterpaper, 10 pt, conference]{ieeeconf}
\IEEEoverridecommandlockouts                                       
 
\usepackage{cite}
\usepackage{algorithm, algpseudocode}  
\usepackage{graphics} 
\usepackage{epsfig} 
\usepackage{amssymb} 
\usepackage{amsmath} 
\usepackage{latexsym}
\usepackage{mathrsfs} %
\usepackage{xcolor}
\usepackage{cases}
\usepackage[normalem]{ulem}
\usepackage{bm}       
\usepackage{graphicx}
\usepackage{verbatim}
\usepackage{balance}
\usepackage{geometry} 
\usepackage{booktabs}

\usepackage{environ}

\NewEnviron{ruledtable}{%
\par\addvspace{2mm}\hrule height 0.03cm 
\begin{table}[!h]\BODY\end{table}
\hrule height 0.03cm \addvspace{2mm}
}

\makeatletter
\let\NAT@parse\undefined
\makeatother

\usepackage[linkbordercolor={1 1 1},citebordercolor={1 1
  1},urlbordercolor={0.0 0.0
  0.0},urlcolor=blue,colorlinks=true,linkcolor=black,citecolor=black]{hyperref}
\usepackage{url}

\newcommand{\algo}{{\sc\textsf{LineCoSpar}}}
\newcommand{\newalgo}{{\sc\textsf{LineCoSparNLP}}}
\newcommand{\prefdata}{{\mathbf{D}}}
\newcommand{\exactions}{{\mathbf{E}}}
\newcommand{\actions}{{\mathbf{A}}}
\newcommand{\itact}{{\bm{a}}}
\newcommand{\prefs}{{\bm{p}}}
\newcommand{\randline}{{\mathbf{L}}}
\newcommand{\curspace}{{\mathbf{S}}}
\newcommand{\numcon}{{v}}

\DeclareMathOperator*{\argmin}{argmin}
\renewcommand{\argmin}{\operatornamewithlimits{argmin}} 
\DeclareMathOperator*{\argmax}{argmax}
\renewcommand{\argmax}{\operatornamewithlimits{argmax}} 
\newcommand{\R}{\mathbb{R}} 
\renewcommand{\P}{\mathcal{P}}
\newcommand{\N}{\mathcal{N}} 

\newcommand{\utilvec}{{\bm{U}}}

\title{\LARGE \bf

Preference-Based Learning for User-Guided  \\ HZD Gait Generation on Bipedal Walking Robots

}

\author{Maegan Tucker$^{1}$, Noel Csomay-Shanklin$^{2}$,  Wen-Loong Ma$^{1}$, and Aaron D. Ames$^{1, 2}$
\thanks{This research was supported by NSF NRI award 1924526 and CMMI award 1923239, NSF Graduate Research Fellowship No. DGE‐1745301, and the Caltech Big Ideas and ZEITLIN Funds.}%
\thanks{$^{1}$Authors are with the Department
of Mechanical and Civil Engineering, California Institute of Technology,
Pasadena, CA 91125.}
\thanks{$^{2}$Authors are with the Department
of Computing and Mathematical Sciences, California Institute of Technology,
Pasadena, CA 91125.}%
}

\begin{document}

\maketitle
\thispagestyle{empty}
\pagestyle{empty}

\begin{abstract}

This paper presents a framework that leverages both control theory and machine learning to obtain stable and robust bipedal locomotion without the need for manual parameter tuning. Traditionally, gaits are generated through trajectory optimization methods and then realized experimentally --- a process that often requires extensive tuning due to differences between the models and hardware. In this work, the process of gait realization via hybrid zero dynamics (HZD) based optimization is formally combined with preference-based learning to systematically realize dynamically stable walking. Importantly, this learning approach does not require a carefully constructed reward function, but instead utilizes human pairwise preferences. The power of the proposed approach is demonstrated through two experiments on a planar biped AMBER-3M: the first with rigid point-feet, and the second with induced model uncertainty through the addition of springs where the added compliance was not accounted for in the gait generation or in the controller. In both experiments, the framework achieves stable, robust, efficient, and natural walking in fewer than 50 iterations with no reliance on a simulation environment. These results demonstrate a promising step in the unification of control theory and learning. 

\end{abstract}



\section{Introduction} \label{sec:intro}

Despite advancements within robotics, realizing dynamic bipedal locomotion on hardware \cite{krotkov2017darpa} remains a benchmark problem across the fields of control, engineering, high-performance computing and machine learning. 
The dynamics and control community has historically approached the challenge of walking from theory applied to real-world platforms, for example Raibert's seminal work on hopping robots \cite{raibert1986legged}. Such theory includes locomotion stability, which has been well studied and realized experimentally from various control perspectives including \textit{zero moment point} (ZMP) \cite{sugihara2002real} and simple model-based methods, such as LIP  \cite{kajita20013d}, SLIP \cite{poulakakis2009spring}, and centroidal dynamics \cite{orin2013centroidal}. These methods, although powerful, do not account for the full-order dynamics of the system. 

Alternatively, the \textit{hybrid zero dynamics} (HZD) framework reduces the full-order dynamics to a lower-dimensional zero dynamics manifold, through which stability of the overall system can be certified. This is accomplished by first characterizing walking as a hybrid system with continuous dynamics and discrete state jumps. The HZD framework then uses Lyapunov methods to guarantee stability of the entire hybrid system \cite{grizzle20103d,ames2014rapidly,ames2016control}. This approach has been demonstrated for walking \cite{sreenath2011compliant,reher2019dynamic,reher2020algorithmic}, running \cite{ma2017bipedal}, and quadrupedal locomotion \cite{ma2019first}.
To accomplish experimental success, however, one needs more than the theoretical stability guarantees --- one must achieve robustness against unmodeled dynamics, which is especially difficult for model-based methods such as the HZD framework.  
%
This ``last-mile mission'' was historically solved by intensive parameter tuning, an arduous and nonintuitive process which inevitably affects the scalability of translating theory to hardware in a practical setting. 

To circumvent this engineering empiricism, the field of machine learning has approached bipedal locomotion from different perspectives, including reinforcement leaning and imitation learning. Reinforcement learning simplifies the process of ``learning to walk'' \cite{learntomove} without prior knowledge \cite{castillo2019reinforcement, hitomi2006reinforcement, ha2020learning, morimoto2004simple}, 
but because this methodology relies on a carefully crafted reward function, the behavior is exclusively determined by its construction. This motivates the second method, imitation learning, which infers the underlying reward function from expert demonstrations \cite{hwangbo2019learning, tirumala2020learning, xie2019iterative}. While both methods have demonstrated promising results, they heavily rely on physical engines such as Bullet \cite{coumans2019}, MuJoCo \cite{todorov2012mujoco}, and RaiSim \cite{RaiSim}. As realistic as these rigid-body-dynamics based simulation environments have become, they still struggle with rough-terrain dynamics such as elastic impacts, slipping contacts, and granular media.
These differences become more apparent when transferred to real-world systems. 
%
%
\begin{figure}[tb]
    \centering
    \vspace{1mm}
    \includegraphics[width=0.98\linewidth]{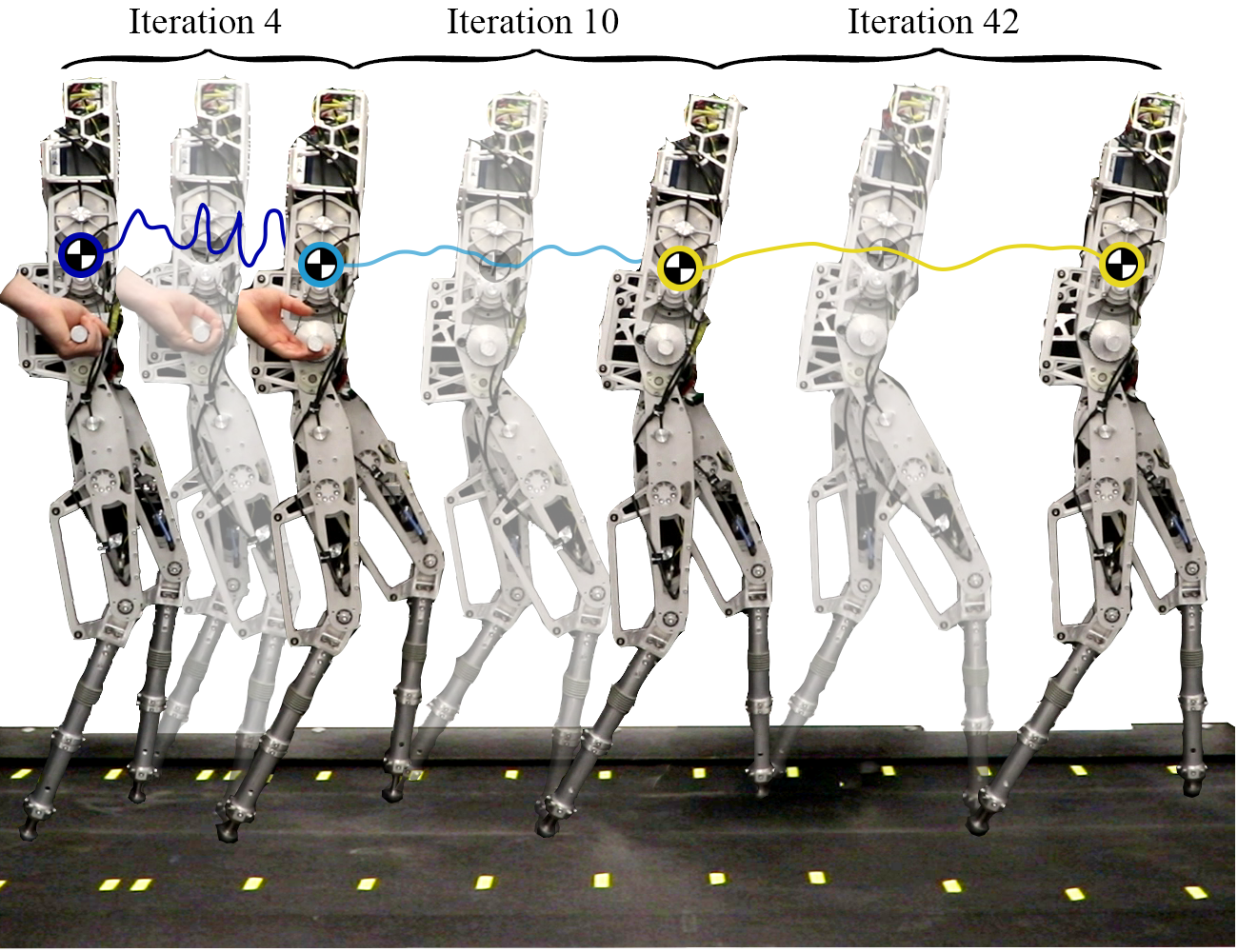}
    \caption{ 
    Through 50 iterations of experiments, the proposed combination of preference-based learning and HZD optimization transforms failed gaits into robust walking on the AMBER-3M robot with a pair of compliant legs.}
    \label{fig:compliantTiles}
\end{figure}
As opposed to relying on just one field, this paper explores combining the successes of both: the formality of stability from control theory and the ability to learn the relationship between complex parameter combinations and their resulting locomotive behavior from machine learning. This is accomplished by building upon our previous results \cite{tucker2020preference, tucker2020human} and systematically integrating preference-based learning with gait generation via HZD optimization. The result is optimal walking on hardware based only on pairwise preferences from the operator (i.e. the user prefers gait A over gait B). We demonstrate the power of this framework through two experiments on a modular biped, AMBER-3M, shown in Fig. \ref{fig:compliantTiles}. In both experiments, stable, robust, efficient, and visually appealing walking is achieved on hardware in fewer than 50 iterations, with no reliance on a simulation environment.
\section{HZD Gait Generation} \label{sec:opt}

The underlying control scheme of the proposed learning framework is based around two concepts: (1) hybrid zero dynamics (HZD) \cite{grizzle20103d,ames2014rapidly}, which theoretically addresses locomotion stability, and (2) trajectory optimization, namely direct collocation \cite{hereid2017frost}, which produces a walking trajectory (gait) that encodes the stability of the closed-loop system. We will briefly review this methodology in this section.


\subsection{Hybrid Zero Dynamics Method} 

Inherently, locomotion consists of alternating sequences of continuous-time dynamics and discrete-time impacts, which can be encoded as a hybrid control system \cite{westervelt2018feedback}. Consider a robotic system with the configuration coordinates $q\in \mathcal{Q}\subset \mathbb{R}^n$ and the full system state $x=(q,\dot{q})\in \mathcal{X}\subset T\mathcal{Q}$.
%
The continuous-time control system is given by:
\begin{align}\label{eq:continuous}
    D(q) \ddot q + H(q,\dot q) = Bu, 
\end{align}
where $D(q)\in\R^{n\times n}$ is the inertia matrix, $H(q,\dot{q})\in\R^n$ is the drift vector, $B\in\R^{n\times m}$ is the actuation matrix, and $u\in\mathcal{U}\subset \R^m$ is the input. 
Here we present the ``pinned'' model for notional simplicity, but the ``unpinned model'' could similarly be considered \cite{hereid2018dynamic}. 
Note that $m<n$ for underactuated robotic systems, such AMBER-3M.

As the robot's foot strikes the ground, an instantaneous change in velocity occurs causing the system state to suddenly jump. Taking $z:\mathcal{Q}\to \mathbb{R}$ to represent the height of the swing foot, the admissible states are given by the \emph{domain}: $\mathcal{D} := \{ (q,\dot{q}) \in \mathcal{X} \ | \ z(q) \geq 0\} \subset \mathcal{X}$.  The region where this instantaneous change in velocity occurs is given by the \textit{switching surface} $\mathcal{S}\subset \mathcal{D}$ defined by: 
\begin{align}
    \mathcal{S} := \{(q,\dot{q}) \in \mathcal{X}\ |\ z(q) = 0, \dot{z}(q,\dot{q}) < 0\}. \label{eq:guard}
\end{align} Taking $x:=(q,\dot{q})$, the discrete dynamics during this impact event are encoded by the \textit{reset map} $\Delta:\mathcal{S}\to \mathcal{X}$, defined as:
\begin{align}
    ~x^+ = \Delta(x^-), \hspace{5mm}x^-\in\mathcal{S} \label{eq:resetMap}
\end{align}
where the $x^+$ and $x^-$ denote the pre- and post-impact state respectively. 
Finally, one can
convert \eqref{eq:continuous} to a \emph{control system}: $\dot x = f(x) + g(x) u$, where when combined with (\ref{eq:guard}) and (\ref{eq:resetMap})
 yields the single-domain hybrid control system:
 \begin{align}
    \mathcal{H}\mathcal{C} =  \begin{cases}{}
         \dot{x} = f(x) + g(x) u   & x \notin \mathcal{S}  \\
        x^+ = \Delta(x^-) & x^- \in \mathcal{S} ,
    \end{cases}
 \end{align}
which can be extended to the multi-domain case; for more details on both single and multi-domain models, refer to \cite{grizzle20103d}. 
%
%
%

The HZD framework reduces the system $\mathcal{HC}$
to a lower-dimensional system. Consider the \textit{zero dynamics surface}:
\begin{align*}
    \mathcal{Z}_\alpha := \{ x \in \mathcal{D}\ |\ &y(q,\alpha) = 0,\ \dot{y}(q,\alpha) = 0 \},
\end{align*}
where $y:\mathcal{Q} \to \mathbb{R}^m$ is defined through the following \emph{outputs} or \emph{virtual constraints} (encoding desired behavior):  
\begin{align} \label{eq:output}
    y(q,\alpha) = y^a(q) - y^d(\tau(q),\alpha).
\end{align} 
Here, $y^a(q)$ is the actual measured output of the system, and $y^d(\tau(q),\alpha)$ is the desired output.  For the following discussion, we take the desired output to be parameterized by the state-based timing variable $\tau(q)$ and a collection of B\'ezier coefficients $\alpha$. Through the use of a stabilizing controller $u^*(x)$, e.g., given by feedback linearization or control Lyapunov functions \cite{westervelt2018feedback,ames2014rapidly,ames2016control}, one can drive $y \to 0$ exponentially. The end result is the \emph{closed-loop dynamics}: $\dot{x} = f_{\rm cl}(x) = f(x) + g(x) u^*(x)$.  In order to guarantee stability of a hybrid system, a hybrid invariance condition must be satisfied, encoded through the \textit{HZD condition}:
\begin{align}
    \Delta(\mathcal{S} \cap \mathcal{Z}_\alpha) \subset \mathcal{Z}_\alpha. \label{eq:HZD}
\end{align}
The remaining step to achieving hybrid invariance is to generate $\alpha$ such that the HZD condition is satisfied.

\subsection{Trajectory Optimization} 
To obtain $\alpha$, we use a direct collocation based optimization algorithm, FROST \cite{hereid2017frost}, which has been previously utilized for efficient gait generation of walking \cite{reher2019dynamic}, running \cite{ma2017bipedal}, and quadrupedal locomotion \cite{ma2020coupled}. Direct collocation is an implicit Runge–Kutta method to approximate the numerical solution of certain dynamical systems, namely differential-algebraic equations and partial differential equations.
The trajectory optimization problem is stated as:

\begin{ruledtable}
\vspace{-2mm}
{\textbf{\normalsize HZD Optimization:}}
\par\vspace{-4mm}{\small 
\begin{align*} \label{eq:opt}
   \{\alpha^*,X^*\} = \argmin_{\alpha,X} &~  \Phi(X) \\
    \text{s.t.}\quad 
    & \dot{x} = f_{cl}(x) \tag{Closed-loop Dynamics} \\
    & \Delta(\mathcal{S} \cap \mathcal{Z}_\alpha) \subset \mathcal{Z}_\alpha \tag{HZD Condition} \\
    & X_{\text{min}}  \preceq X \preceq X_{\text{max}} \tag{Decision Variables} \\
    & c_{\text{min}}  \preceq c(X) \preceq c_{\text{max}} \tag{Physical Constraints} \\
    & a_{\text{min}} \preceq p(X) \preceq a_{\text{max}} \tag{Essential Constraints} \label{eq: essential} 
\end{align*}}\vspace{-6mm}\par
\end{ruledtable}
\noindent where $X = (x_0,...,x_N,T)$ is the collection of all decision variables with $x_i$ the state at the $i^{th}$ discretization and $T$ the duration, $\Phi(X)$ is the cost function, and $c(X)$ is the set of physical constraints on the optimization problem. These physical constraints are included in every gait generation framework to encode the physical laws of real-word, such as the friction cone condition, workspace limit, and motor capacity \cite{reher2020algorithmic}. In this work, we specify a specific subset of physical constraints as $p(X)$, which we term \textit{essential constraints} and discuss further in Sec. \ref{sec:essential}.
With this optimization formulation, we can use nonlinear programming (NLP) solvers, such as IPOPT \cite{wachter2006implementation}, to efficiently synthesize an optimal walking gait. The end result is a stable periodic solution to the walking dynamics that is parameterized by some static set of B\'ezier coefficients $\alpha^*$. 

\subsection{Essential Constraints} 
\label{sec:essential}
Expert operators typically tune $a_{\text{min}} \in \R^{\numcon}$ and $a_{\text{max}} \in \R^{\numcon}$ of \eqref{eq: essential} in the hopes of guiding the HZD optimization towards a solution that maximizes the operators' subjective metric of ``good'' walking. Since the construction of these constraints is often essential towards achieving experimental robustness, we term them \textit{essential constraints}. Traditionally, essential constraints consist of gait features such as average velocity, step length, foot clearance, and impact velocity.
%
Often, practitioners derive intuition on how to shape essential constraints from years of experience.
One example of how this intuition relates to stability is
Raibert-type controllers \cite{raibert1986legged}, which tune the relationship between step length and walking velocity based on a simplified model.

In this paper, we present a systematic approach towards tuning essential constraints using preference-based learning. To do so, we reformulate \eqref{eq: essential} as:
\begin{align*}
    a - \delta \preceq p(X) \preceq a + \delta,
\end{align*}
where $a \in \R^{\numcon}$ consists of $\numcon$ constraint values, and $\delta \in \R^{\numcon}$ defines the equality tolerance for each constraint. Thus, the goal of the learning is to identify $a^* := \argmax_{a \in \R^{\numcon}} ~ U(a)$, where $U: \R^{\numcon} \to \R$ is the underlying utility function. In our work, we construct the components of $a$ to be: 
\begin{enumerate}
    \item average forward velocity of the torso (m/s)
    \item phase variable value at which to enforce minimum foot clearance, $\tau_{c}$
    \item minimum nonstance foot clearance enforced at $\tau_{c}$ (m)
    \item downward velocity enforced at impact (m/s)
    \item step length, i.e. the forward distance between swing foot and stance foot at impact (m),
\end{enumerate}
which are defined over the search space of possible parameter combinations $\actions$, a discretization of $\R^{\numcon}$, as given in Table \ref{table:actionbounds}.

\subsection{Benefits of Preference-Based Learning}
The traditional hand-tuning process requires a human operator to make assumptions about the underlying utility function $U$, which is difficult given the following: the non-intuitive relationship between parameter combinations and the resulting experimental behavior; and the need to account for numerous factors including stability, robustness to perturbations/model uncertainty, and visual appearance. Additionally, $U$ admits no obvious mathematical description; eliminating the use of reward-based tuning methods. 

Alternatively, we propose the use of preference-based learning to identify $a^*$ using only pairwise preferences, which take advantage of a human's natural ability to combine many factors into a single judgment of ``better'' or ``worse''. 
Although this requires the human to provide feedback, there are two major benefits of our approach: 1) the duration of the tuning process is reduced significantly compared to hand-tuning; and 2) pairwise preferences are much easier for a na\"ive user to provide compared to manually navigating the complex search space of parameter combinations.


\begin{table}[tb]
    \centering
    \caption{Essential Constraint Action Space}
    \begin{tabular}{|c|c|c|}
        \hline
        Essential Constraint & Bounds [$a_{\text{min}}, a_{\text{max}}$] & Disc. $d$ \\
        \hline
        Average Forward Velocity (m/s) & $[0.3,0.6]$ & $0.05$ \\
        \hline
        Clearance Tau $(\cdot)$ & $[0.4, 0.7]$ & $0.1$ \\
        \hline
        Minimum Foot Clearance (m) & $[0.05, 0.19]$ & $0.02$ \\
        \hline
        Impact Velocity (m/s) & $[-0.8, -0.2]$ & $0.1$ \\
        \hline
        Step Length (m) & $[0.2,0.4]$ & $0.05$ \\
        \hline
    \end{tabular}
    \label{table:actionbounds}
\end{table}

%
%
%
%


\section{Learning Framework} \label{sec:framework}
To learn the optimal action $a^*$ in as few iterations as possible, we introduce a framework built around a high-dimensional preference-based learning algorithm \algo \cite{tucker2020human} that learns a Bayesian posterior over the utility function $U$. The new framework, \newalgo, still relies on pairwise preferences obtained from a human observing the experimental behavior of the robot, but embeds the learning directly into an HZD optimization problem, eliminating the need for a pre-computed gait library. We will first present \newalgo, and then explicitly discuss the differences between the two frameworks.

\begin{algorithm}[tb]
\caption{\newalgo}
\begin{small}
\begin{algorithmic}[1]
\State Construct $\actions$ using $a_{\text{min}}$, $a_{\text{max}}$, and $d$
\State Initialize datasets \{$\prefdata_0, \exactions_0 = \emptyset \} $






\ForAll {$i = 1,\dots, N$}{}


    \If{ $i == 1$}
        \State Obtain $\itact_1 = \{ a_{1}^1, ...., a_{1}^n \}$ as uniform-random
    \Else
        \State Generate $\randline_{i} := $ random line through $a^*_{i-1}$
        \State Construct subset $\curspace_i = \randline_i \cup \exactions_{i-1}$ 
        \State Approximate $\P(\utilvec_{\curspace_i} | \prefdata_{i-1})$ as $\N(\mu_{\curspace_i},\Sigma_{\curspace_i})$
        \State Draw $k = 1,...,n$ samples: $f^k \sim \mathcal{N}(\mu_{\curspace_i}, \Sigma_{\curspace_i})$
        \State Obtain $\itact_i = \{ a_{i}^k = \underset{{a\in \curspace_i}}{\text{argmax}} f^k(a) | k = 1,...n \}$
    \EndIf


\State Execute outputs of NLP for $\itact_i$ on the system
\State Append executed actions:  $\exactions_{i} = \exactions_{i-1} \cup \itact_i$
\State Query operator for preference feedback $\prefs_i$ 
\State Append preference feedback: $\prefdata_{i} = \prefdata_{i-1} \cup \prefs_i$

\State Approximate $\P( \utilvec_{\exactions_i}| \prefdata_{i})$ as $\N(\mu_{\exactions_i},\Sigma_{\exactions_i})$

\State Update $a^*_{i} = \underset{a \in \exactions_{i}}{\text{argmax}} ~ \mu_{\exactions_i}(a)$

\EndFor
\end{algorithmic}
\end{small}
 \label{alg:linecospar}
 \end{algorithm}

\subsection{The \newalgo~ Algorithm}
The procedure of the \newalgo~ algorithm is shown in Alg. \ref{alg:linecospar}. First, to set up the learning problem, upper and lower bounds on $a \in \R^{\numcon}$ along with the granularity of discretization $d \in \R_+^{\numcon}$ are chosen by the operator. This leads to the discrete search space $\actions$ with $|\actions| = \prod d$. The corresponding set of utilities is defined as $\utilvec: \actions\to \R$, with $\utilvec_{B}$ used to denote the restriction of $\utilvec$ on $B \subset \actions$.

 


Each iteration $i$ of the algorithm is as follows. First, $n$ actions, denoted as the set $\itact_i:= \{a_i^1, \dots, a_i^n\} \in \R^{\numcon \times n}$, must be selected to give to the NLP. The parameter $n$ can be changed depending on how many actions the operator would like to sample in each iteration. Because the actions are compared in pairs, $n$ actions equates to $m=\binom{n}{2}$ pairwise preferences. In the first iteration, $\itact_1$ is constructed using uniform-random actions. During every subsequent iteration, the algorithm utilizes a Self-Sparring approach \cite{sui2017multi} to Thompson sampling which is a sample-efficient sampling method for regret-minimization. In general, to select $n$ actions, Thompson sampling works by drawing $n$ samples from a given distribution, such as the normal distribution $\N(\mu_B, \Sigma_B)$ over actions $a \in B \subset \actions$:
\begin{align}
    f^k \sim \N(\mu_B, \Sigma_B) \quad \forall k = 1,\dots,n, \label{eq: Thompson1}
\end{align}
and selecting the actions that maximize the samples:
\begin{align}
    a_{i}^k = \argmax_{a \in B} f^k(a) \quad \forall k = 1,\dots,n. \label{eq: Thompson2}
\end{align}

\noindent To be computationally tractable, \newalgo~ performs Thompson sampling only considering the subset of actions $\curspace_i \subset \actions$. This subset is defined as $\curspace_i := \randline_i \cup \exactions_{i-1}$, with $\exactions_{i-1}$ being the dataset of previously executed actions and $\randline_i\subset \actions$ being a random linear subspace which intersects the best action from the previous iteration, $a^*_{i-1}$. Using this subset, Thompson sampling draws $n$ samples from the posterior distribution $\P(\utilvec_{\curspace_i}|\prefdata_{i-1})$, where $\prefdata_{i-1}$ is the preference dataset from the previous iteration. The posterior is modeled as proportional to the product of the preference likelihood and the Gaussian prior \cite{chu2005preference}:
\begin{align}
    \P(\utilvec_{\curspace_i}|\prefdata_{i-1}) \propto \P(\prefdata_{i-1}|\utilvec_{\curspace_i})\P(\utilvec_{\curspace_i}).
    \label{eq: posterior}
\end{align}
The Gaussian process prior is computed as:
\par\vspace{-3mm}{\small 
\begin{align}
    \P(\utilvec_{\curspace_i}) = &\frac{\exp \left( -\frac{1}{2} \utilvec_{\curspace_i} (\Sigma_i^{\text{pr}})^{-1} \utilvec_{\curspace_i} \right)}{(2\pi)^{\frac{|\curspace_i|}{2}}|\Sigma_i^{\text{pr}}|^{1/2}}, \label{eq: prior}
\end{align}}
where $\Sigma_i^{\text{pr}} \in \mathbb{R}^{|\curspace_{i}| \times |\curspace_{i}|}$ with $[\Sigma_i^{\text{pr}}]_{j,k} = \mathcal{K}(a_{\curspace_i}^{j}, a_{\curspace_i}^{k})$ for the set of actions $a_{\curspace_i}$ in $\curspace_i$, and $\mathcal{K}$ being a kernel of choice (taken as a squared exponential kernel in this work). The preference likelihood function is computed as:
\par\vspace{-3mm}{\small 
\begin{align}
    \P(\prefdata_{i-1} | \utilvec_{\curspace_{i}}) = \prod_{j=1}^{i-1}\prod_{k=1}^{n} g \left( \frac{U(a_{j}^k) - U(a_{j}^k)}{c_p}\right), \label{eq: likelihood}
\end{align}}\vspace{-3mm}\par\noindent
where $g : \R \to (0,1)$ is a monotonically-increasing activation function, and $c_p > 0$ models the expected noisiness of the preference feedback. In this work, we select $g(x) := \frac{1}{1+e^{-x}}$ to be the heavy-tailed sigmoid function because it was empirically found to improve performance \cite{tucker2020human}. 
 
Equipped with \eqref{eq: prior} and \eqref{eq: likelihood}, the posterior \eqref{eq: posterior} can then be estimated via the Laplace approximation as in \cite{chu2005preference} which yields a multivariate Gaussian, $\mathcal{N}(\mu_{\curspace_i}, \Sigma_  {\curspace_i})$. Finally, applying this distribution to \eqref{eq: Thompson1} and \eqref{eq: Thompson2} yields $\itact_i$. These sampled actions are then given to the NLP, whereby corresponding gaits are generated, the outputs are executed on the robot, and $\itact_i$ is appended to $\exactions_{i}$. We define the set of actions executed on hardware up to and including those sampled in iteration $i$ as $\exactions_{i} := \{\itact_1, \dots, \itact_i\} \in \R^{v \times n \times i } \subset \actions$.

After demonstrating the gaits on hardware, the human operator is queried for $m$ pairwise preferences, denoted as $\prefs_i = \{p_{i}^1, \dots, p_{i}^m\}\in \R^{m}$. The collection of all preference feedback up to and including iteration $i$ is denoted $\prefdata_i := \{\prefs_1, \dots, \prefs_i\} \in \R^{m \times i}$. Note that it is possible for $\prefs_i = \emptyset$ when all sampled actions do not converge, or when the user chooses to give feedback of ``no preference''. 

Lastly, the algorithm updates its belief of $a^*$ by modeling the posterior again using $\prefdata_i$. Since obtaining the posterior over the entire search space $\actions$ for high-dimensional action spaces has been shown to be computationally intractable \cite{tucker2020human}, the posterior is only updated over $\exactions_i$:
\begin{align}
    \P(\utilvec_{\exactions_{i}}|\prefdata_{i}) \propto \P(\prefdata_{i}|\utilvec_{\exactions_{i}})\P(\utilvec_{\exactions_{i}}), 
\end{align}
which is approximated using the same procedure as for $\P(\utilvec_{\curspace_{i}}|\prefdata_{i-1})$ and applying the Laplace approximation to obtain the distribution $\N(\mu_{\exactions_i}, \Sigma_{\exactions_i})$. The algorithm's belief of the optimal action after iteration $i$ is finally updated as: 
\begin{align*}
    {a}^*_{i} = \underset{a \in \exactions_{i}}{\text{argmax}} ~\mu_{\exactions_i}(a).
\end{align*}


\subsection{Changes to \algo~ for use with a NLP}
Three notable changes were made to the algorithm \newalgo~ in comparison to \algo. First, the \newalgo~ selects $\randline_i$ to intersect $a^*_{i-1}$ as opposed to $a^*_{i-2}$ which leverages more recent preference feedback. This change requires two posterior updates in each iteration but results in fewer required iterations. Second, \algo~ uses a buffer method to compare executed actions with previously executed actions which results in higher sample-efficiency. However, when considering preference-based learning towards gait generation, it is important to account for the computation time required to obtain gaits. For this reason, we modify the \newalgo~ algorithm to sample and query $n>1$ actions in each iteration. This results in worse sample-efficiency, but allows for batched gait generation that enables the generated gaits to be executed on hardware back to back. Lastly, in \algo, coactive feedback, otherwise known as user suggestions, is also added to the dataset $\prefdata_{i}$ to improve sample-efficiency. However, these suggestions rely on understanding the mapping between $a$ and $U(a)$; because this mapping is rarely well-understood for parameters of a nonlinear optimization problem, \newalgo~ does not utilize coactive feedback.


\begin{figure}[tb]
    \centering
     \includegraphics[width=0.45\textwidth]{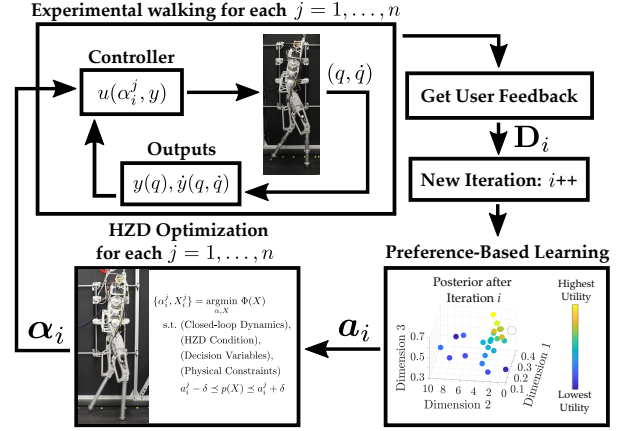}
    \caption{The experimental procedure is illustrated in terms of each iteration $i$ with $n$ denoting the number of gaits compared in each iteration. The experiments presented in this work used $n = 2$. Using this notation, the set of $n$ actions given to the HZD optimization is denoted: $\itact_i = \{ a_{i}^1, \dots, a_{i}^n \}$. The resulting $n$ sets of B\'ezier coefficients given to the controller are denoted $ \bm{\alpha}_i = \{ \alpha_{i}^1, \dots, \alpha_{i}^n \} $.}
    \label{fig:expProcedure}
\end{figure}

\section{Learning to Walk in Experiments} \label{sec:exp}

We experimentally deploy \newalgo~ (open-source code: \cite{repository}) to tune the 5 essential constraints outlined in Table \ref{table:actionbounds} on the planar bipedal robot, AMBER-3M \cite{Ambrose17toward}. This custom research platform has three interchangeable lower-limb configurations: flat-foot, point-foot, and spring-foot. We specifically selected this platform because of its engineering reliability \cite{ma2019dynamic}, enabling consistent data collection to isolate the effects of various gaits in the learning process. The controller for AMBER-3M is implemented on an off-board i7-6700HQ CPU @ 2.6GHz with 16 GB RAM, which computes desired torques and communicates them with the motor drivers. The motor driver communication and the control logic run at $\sim$1kHz, each on a separate core.

\subsection{Experimental Procedure}
In the experiments, walking gaits are generated by the HZD-based method presented in Sec. \ref{sec:opt}. We take $y^a(q) := q^a\in\R^4$ as the position of the four motorized joints of AMBER-3M, $\tau(q)$ to be the linearized forward hip position, and use a $5^{th}$-order B\'ezeir polynomial ($\alpha\in\R^{4\times 6}$) to describe the desired output trajectories. Additionally, the cost function is selected to be the mechanical cost of transport (MCOT), a common metric for locomotion efficiency:
\begin{align}
    MCOT = \int_{t_0}^{t_f} \frac{P(t)}{mgv}dt, \label{eq:MCOT}
\end{align}
where $P(t) = \sum_{i=1}^4 |u_i(t) \dot{q}^a_i(t) |$ is the 2-norm sum of power. 

The average optimization run time is 0.1 second per iteration, with each gait averaging 160 iterations. The experimental procedure is illustrated in Fig. \ref{fig:expProcedure}. In our experiments, the learning was conducted for $n=2$, corresponding to two gaits being compared in each iteration. This was chosen because we empirically found that operators sometimes had difficulty remembering the details of more than two gaits at a time, leading to the most reliable preference feedback when $n=2$. Note that other applications may benefit in a higher $n$, which would increase the rate of learning.

Each trial began by initializing AMBER-3M in a static double-support configuration, starting the treadmill, and attempting to push the robot into the designed periodic orbit. If the resultant dynamics were not stable, extra precaution was taken to give the gait the best chance at succeeding. Once the gait reached its orbit, the robot was released and the robustness of the gait to various disturbances was investigated. After both gaits were executed on the physical robot, a preference was collected from the human operator observing the physical realization of the walking. In some iterations, video footage was also reviewed before giving a preference. The criteria used to determine preferences between gaits were the following (in order of prioritization):
\begin{itemize}
    \item Capable of walking
    \item Robust to perturbations in treadmill speeds
    \item Robust to external disturbance
    \item Does not exhibit harsh noise (e.g. during impact)
    \item Is visually appealing (intuitive judgment from operator)
\end{itemize}

\begin{figure}[tb]
    \centering
    \includegraphics[width=\linewidth]{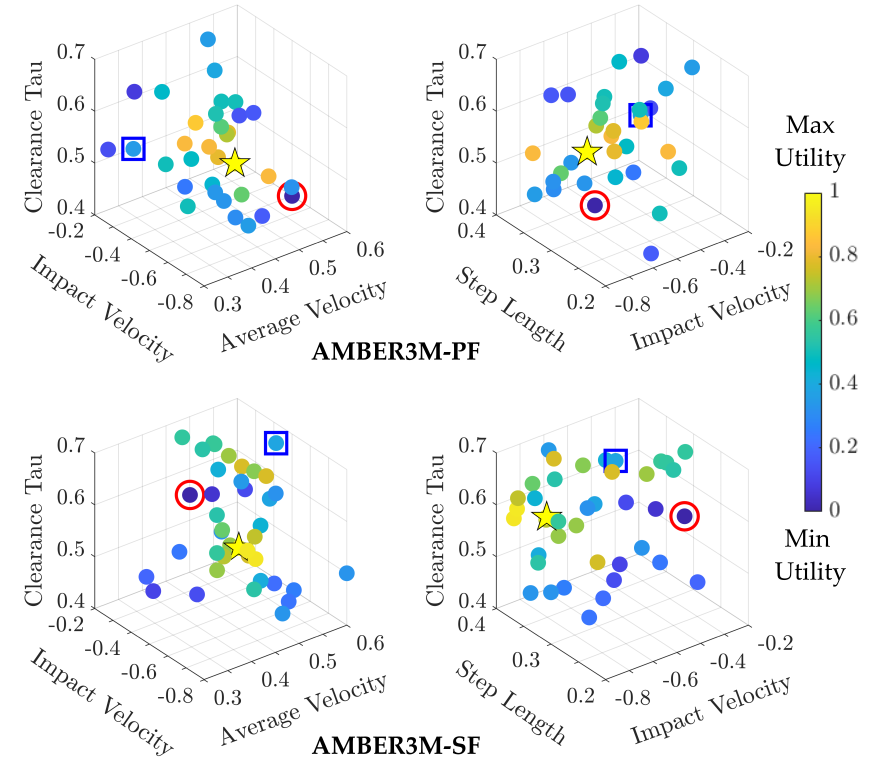}
    \caption{The final obtained utilities for the visited actions, averaged over the two dimensions not shown on each subplot.
    The optimal action is illustrated by the yellow star ($[0.4399,0.5425, 0.0759, -0.6040, 0.3190]$ for AMBER3M-PF and $[0.4105, 0.5930, 0.0833, -0.7020, 0.3504]$ for AMBER3M-SF). The other two actions depicted in Fig. \ref{fig:plots} are denoted with a red circle (worst gait) and a blue square (middle gait).}
    \label{fig:posteriors}
\end{figure}

\begin{figure*}[tb]
    \includegraphics[width=0.99\linewidth]{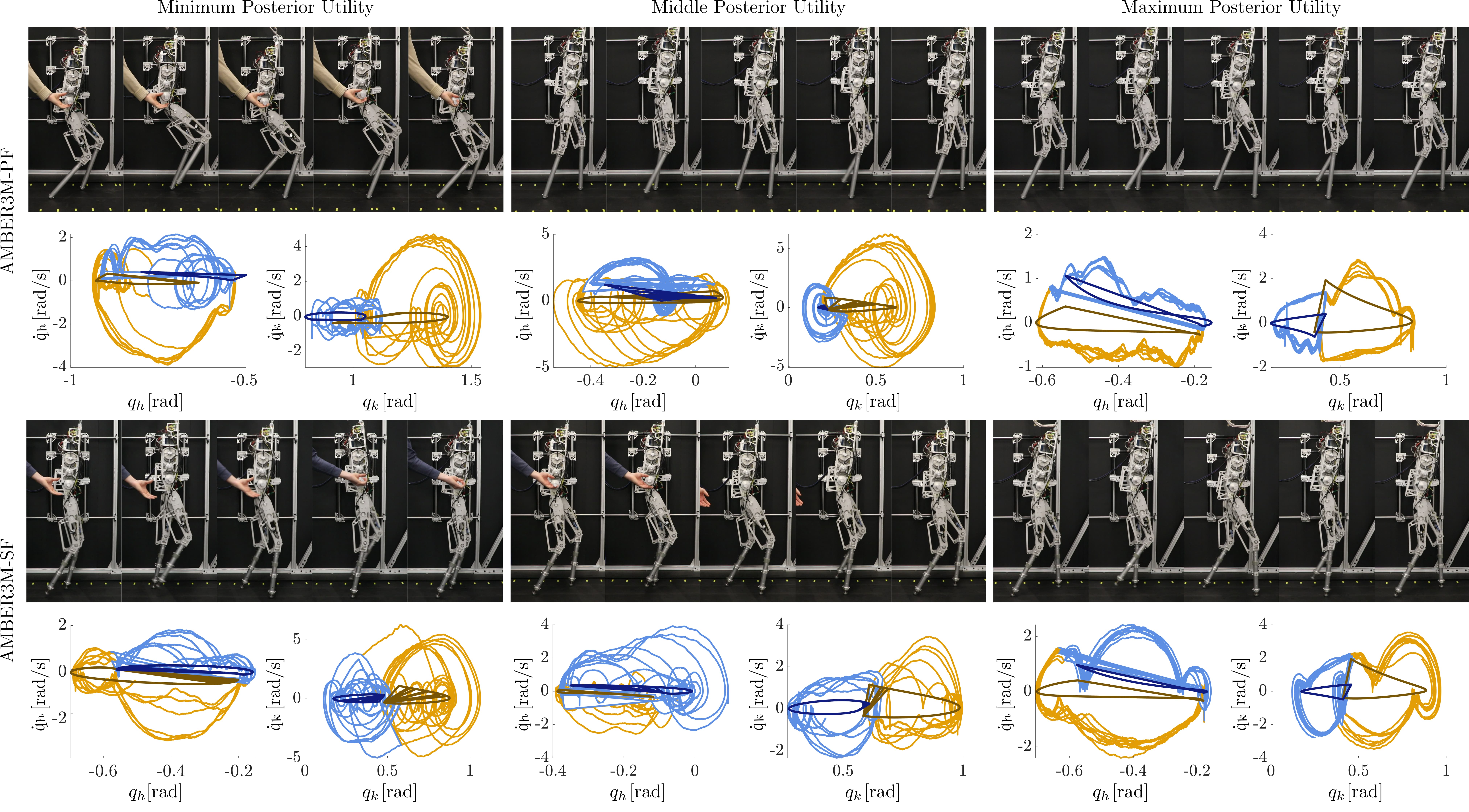}%
    \caption{Gait tiles with increasing posterior utility values from left to right are shown for the the rigid model (top) and spring model (bottom). The phase portraits of the hip ($q_h$) and knee ($q_k$) of the stance leg (blue) and swing leg (yellow) are shown below each corresponding gait, plotted over 10 seconds of data.
    The phase portraits clearly indicate that for both AMBER3M-PF and AMBER3M-SF the gaits evolved to be more experimentally robust.}
    \label{fig:plots}
    \vspace{-10pt}
\end{figure*}

\subsection{Procedure specific to AMBER3M-PF and AMBER3M-SF}
\label{sec:compliance}
In this work, we leverage two configurations of the robot: 1) the point-foot configuration, AMBER3M-PF (1.373 m, 21.3 kg); and 2) the spring-foot configuration, AMBER3M-SF (1.430 m, 23.5 kg) \cite{Ambrose17toward}. We first demonstrate the learning framework on AMBER3M-PF, with the corresponding rigid point-foot model used in the gait generation. To emphasize the scalability of our method, we repeat the exact procedure applied to AMBER3M-PF on AMBER3M-SF, but intentionally do not account for changes in the robot model and instead still generate gaits assuming the rigid-body model. Furthermore, we execute the gaits on hardware using the same controller with unmodified gains. Historically, robots with compliance are difficult to generate gaits for because of the resulting complexities which include: increased degrees of freedom of the system; the addition of a double support domain to the hybrid dynamics; and increased stiffness of the dynamics. Past success with compliant bipeds has relied on sophisticated models \cite{hereid2014dynamic}. Therefore, the fact that our method yields stable walking despite the unmodeled compliance highlights it's effectiveness.



%
%

\subsection{Results}
A summary of the experimental results is illustrated in the supplementary video \cite{video}, with additional videos and material available at \cite{website}, and the final obtained posterior provided with the framework code in the repository \cite{repository}. 

The experiment with AMBER3M-PF was run for 30 iterations and sampled 27 unique gaits. The final posterior over the 27 executed actions is illustrated in the top row of Fig. \ref{fig:posteriors}. Since gaits quickly met the first criterion of being able to walk, preferences were mainly dictated based on the robustness and appearance of the experimental walking. The initial gaits tried on hardware, although optimal subject to the imposed constraints, resulted in inferior trajectory tracking and power consumption. As the algorithm progressed, the gaits became significantly smoother, more robust to disturbance, and energy efficient. This is exemplified in Fig. \ref{fig:plots} which illustrates the gaits corresponding to the minimum, a middle, and the maximum posterior utility; the iterations corresponding to when these gaits were first sampled is 1, 21, and 26, respectively. In Fig. \ref{fig:plots}, we note significantly lower velocity overshoot for all of the limbs and tighter tracking shown in the phase portraits for the gaits with higher posterior utility. It is also interesting to note the framework's success at improving the efficiency of the experimental walking: a latent property which is discernible to the human operator even though it is not immediately measured. This improvement is demonstrated by the MCOT values of the three gaits in Fig. \ref{fig:plots}: 0.74, 0.95, and 0.26 respectively.

When the procedure was repeated on AMBER3M-SF, many of the initial gaits were unable to walk due to the unmodeled compliance. Thus, gaits exhibiting periodic walking were strongly preferred. This second experiment was conducted for 50 iterations and sampled 37 unique gaits with the obtained posterior illustrated in the bottom row of Fig. \ref{fig:posteriors}. Again, three gaits are selected for further discussion corresponding to the minimum, a middle, and the maximum posterior utility values. Gait tiles and phase portraits for these are again shown in Fig. \ref{fig:plots}. The iterations when these gaits were first sampled are 4, 10, and 42. Once again, the algorithm converges to gaits with superior trajectory tracking and lower MCOT (1.16, 0.38, and 0.33, respectively).




\vspace{-2mm}
\section{Conclusion} \label{sec:conc}
\vspace{-2mm}
In this work, we present and experimentally demonstrate a high-dimensional preference-based learning framework, \newalgo~ (open-source code: \cite{repository}), specifically designed for use towards HZD-based gait generation. \newalgo~ incorporates preference-based learning with an HZD optimization problem to leverage the theoretical benefits of HZD without the challenge of parameter tuning. Furthermore, preference-based learning is a sample-efficient learning method that does not require the user to mathematically define a metric for ``good'' walking. Instead, the framework  relies on easy to provide pairwise preferences. 

The success of the proposed method is demonstrated through its ability to experimentally realize gaits that are stable, robust to model uncertainty, robust to external perturbations, efficient, and natural looking within 50 experimental iterations, with no requirement for simulation. Furthermore, \newalgo~ achieves robust walking with unmodeled compliant legs, a challenging control task which historically relied on sophisticated models.

Future work includes extending this framework to more robotic platforms, such as quadrupeds and 3D bipedal robots, as well as improving the sample-efficiency of the framework through additional qualitative feedback mechanisms such as ordinal labels \cite{chu2005gaussian}. The experimental results presented in this paper demonstrate the rich potential lying in the boundary between machine learning and control theory. It is well-known that control theory provides necessary structure to bipedal platforms, but machine learning can play a critical role in shaping the final behavior of the system. 






\bibliographystyle{IEEEtran}
\balance
\bibliography{IEEEabrv,References}

\end{document}